*Article*

# Accurate Prediction Using Triangular Type-2 Fuzzy Linear Regression

**Assef Zare** [1], **Afshin Shoeibi** [2], **Narges Shafaei** [1,*], **Parisa Moridian** [3], **Roohallah Alizadehsani** [4], **Majid Halaji**[5], **Abbas Khosravi** [4]

[1] Faculty of Electrical Engineering, Gonabad Branch, Islamic Azad University, Gonabad, Iran;
[2] Faculty of Electrical Engineering, Clinical Studies Lab (CSL), K. N. Toosi University of Technology, Tehran, Iran.
[3] Faculty of Engineering, Science and Research Branch, Islamic Azad University, Tehran, Iran.
[4] Institute for Intelligent Systems Research and Innovation (IISRI), Deakin University, Geelong, Australia.
[5] Faculty of Electrical Engineering, Neyshabure Branch, Islamic Azad University, Neyshabure, Iran.

Corresponding authors: Narges Shafaei (narges.shafaei@gmail.com)

**Abstract:** Many works have been done to handle the uncertainties in the data using type 1 fuzzy regression. Few type 2 fuzzy regression works used interval type 2 for indeterminate modeling using type 1 fuzzy membership. The current survey proposes a triangular type-2 fuzzy regression (TT2FR) model to ameliorate the efficiency of the model by handling the uncertainty in the data. The triangular secondary membership function is used instead of widely used interval type models. In the proposed model, vagueness in primary and secondary fuzzy sets is minimized and also, a specified x-plane of observed value is included in the same $\alpha -$ plane of the predicted value. Complex calculations of the type-2 fuzzy (T2F) model are simplified by reducing three dimensional type-2 fuzzy set (3DT2FS) into two dimensional interval type-2 fuzzy (2DIT2F) models. The current survey presents a new regression model of T2F by considering the more general form of T2F membership functions and thus avoids high complexity. The performance of the developed model is evaluated using the TAIEX and COVID-19 forecasting datasets. Our developed model reached the highest performance as compared to the other state-of-art techniques. Our developed method is ready to be tested with more uncertain data and has the potential to use to predict the weather and stock prediction.

## 1. Introduction

Real-life information usually has vagueness; consequently, Zadeh introduced type-1 fuzzy set (T1FS) to model this vagueness [1]. Type 2 fuzzy systems (T2FS) were introduced since T1FS cannot model that information with high-level uncertainty [2]. Despite the ability of T2F in modeling natural phenomena, they are not considered in the beginning due to their high computational complexity. Interval type-2 fuzzy sets (IT2FS) were employed more than other T2F models because of its abilities in modeling phenomena and also its simplicity of calculations. Nevertheless, it is not easy to explain the application of anyT2FS. However, T2FS has a various applications in electrical energy, business and finance, healthcare, automatic control and medical applications [3-6].

Finding a relation between two or more variables is necessary for many applications. Regression analysis is a statistical gadget for discovering these relations. Since people use linguistic terms to judge and evaluate things in their life, fuzzy regression models are suitable for these applications.

There are two types of fuzzy regression (FR) models. The first type is the fuzzy linear regression (FLR) introduced by Tanaka [7-9]. Instead of finding fuzzy coefficients in regression models, this model solves a linear programming problem. The second type is based on the least square error method [10].

Many real-life phenomena can use the fuzzy regression analysis [11-17]. Many works have been done on fuzzy regression models because fuzzy regression has many applications [18-21][2]. Most previous research has been done on type-1 fuzzy models. Considering that the fuzzy type-2 models the ambiguity and uncertainty of real-world phenomena better, it is more reasonable to conduct studies on fuzzy type-2 regression. Nevertheless, those variables are difficult to cope with their given three dimensional (3D) features. It is essential to present simpler, more comprehensible, and practical models of type-2 models.

Few researchers have applied type-2 fuzzy linear regressions (T2FLR) to model the issues [22-24][14-15][25-26]. A T2F qualitative regression technique was introduced by Wei and Watada. They employed a general type-2 fuzzy (GT2F) number, whereas they employed interval type-2. Poleshchuk et al. posited a model for IT2FS according to the least-squares approximation [25]. Hosseinzadeh et al. proposed a model with certain input and output in accordance with weighted goal programming (WGP) [27]. However, in this model, only a few points of fuzzy functions converge.

Few works have focused on the type-2 fuzzy c-regression (T2FCR) models, which are different from type-2 fuzzy regression (T2FR) [28-29]. None of the previous models are able to provide a good model of T2FR, and all of them have reduced the T2FR model to several points of the membership function closer together. An IT2FR model was suggested by Shafaei et al., based on $\alpha$- cuts concepts, reaching good results [30].

In all previous T2FR models, the secondary membership function of T2F membership functions is considered Interval fuzzy set. In order to be able to use the fuzzy type-2 in modeling uncertainties in a more comprehensive way, it is better if the membership functions have a more general shape. Hence, the triangular secondary membership functions are considered. The present work has modeled the uncertainties of the membership function of aT1FS in better way.

Two purposes are met in this paper. The first goal is to present a new T2FR model with triangular secondary membership functions that does not create much computational complexity. The second goal is to show the ability of the proposed model to solve real problems.

The proposed model is a T2FR model established by $\alpha-$ plane concepts, which has shown its ability to forecast the Taiwan Stock Exchange Capitalization Weighted Stock Index (TAIEX). This model reduces the 3D features of T2FS toIT2FS. IT2FS can be fully described by a footprint of uncertainty (FOU). The FOU is created with two fuzzy membership functions of type-1, LMF, and UMF. Therefore, the T2FR model is made by developing these two type-1 fuzzy functions.

One of the primary advantages of this model has more straightforward calculation and imaginable concepts in addition to its superior accuracy. This intelligible model helps researchers to use our developed T2FLR model is other applications like weather and stock value predictions.

## 2. Review of Fuzzy Techniques

**Definition 1**: The principal membership function (PrMF) described as $\mu_A = \int_{x \in X} u/x$, where $f_x(u) = 1$.

**Definition 2**: A T1FS $\tilde{A}$ is normal if : $\sup \mu_{\tilde{A}}(x) = 1$

**Definition 3**: If its UMF is normal, An IT2FS $\tilde{A}$ will be normal (sup $\bar{\mu}_{\tilde{A}}(x) = 1$).

**Definition 4**: A perfectly normal IT2FS$\tilde{A}$, can be provided when both its UMF and LMF are normal.

**Definition 5**: T2FS can be normal provided that it has a PrMF and FOU is a normal IT2FS.

**Definition 6**: T2FS is perfectly normal where it has a normal PrMF and its FOU is a perfectly normal IT2FS.

**Definition 7**: An $\alpha-$plane of a T2FS, $\tilde{A}$, denoted by, $\tilde{A}_{\tilde{\alpha}}$, is the union of the primary memberships of $\tilde{A}$, that secondary grades are equal or higher than $\tilde{\alpha}$ (Figure 1) [18]. $A_\alpha^V(\underline{A}_\alpha, \overline{A}_\alpha)$, where $\mu_{A_\alpha^V}(x) = [\mu_{\underline{A}_\alpha}(x), \mu_{\bar{A}_\alpha}(x)]$ (Figure 2) [31].

**Definition 8:** The $\alpha-$cuts of an IT2FS, $A^V$ is a non-fuzzy set described as follows: $A_\alpha^V(\underline{A}_\alpha, \overline{A}_\alpha)$ Here, $\mu_{A_\alpha^V}(x) = [\mu_{\underline{A}_\alpha}(x), \mu_{\bar{A}_\alpha}(x)]$ (Figures 2 and 3) [31].

Regarding Mencattini et al. [22], this kind of IT2FS can deduced as $\tilde{A}_i = [[\underline{a}_i, \overline{a}_i], b_i, [\underline{c}_i, \overline{c}_i]]$ Such that $LMF_{\tilde{A}_i} = [\overline{a}_i, b_i, \underline{c}_i]$, $UMF_{\tilde{A}_i} = [\underline{a}_i, b_i, \overline{c}_i]$ and $\underline{a}_i \leq \overline{a}_i \leq b_i \leq \underline{c}_i \leq \overline{c}_i$. Figure 3 shows a perfectly normal triangular T2FS that is employed in this paper. The higher the value of the second-order membership function, the thicker the shadow.

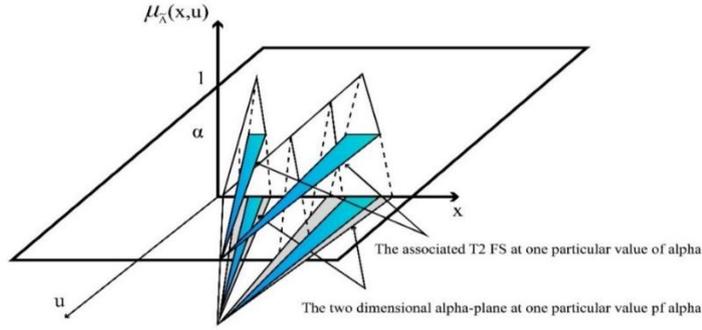

Figure 1. $\alpha$ −Plane representation of T2FSs

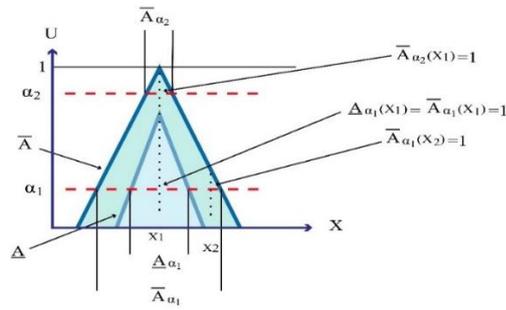

Figure 2. $\alpha$ −Cuts of IVFS

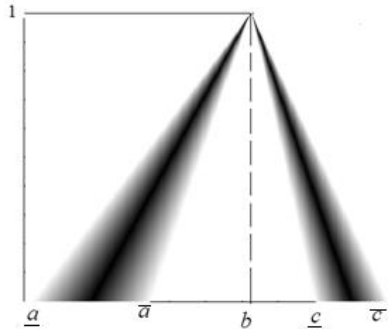

Figure 3. A Perfectly Normal triangular T2FS

## 3. Method

### 3.1. Interval Type-2 Regression Model

For the proposed T2FR model, IT2FR model have been used. Many IT2FR models [32][25] have been proposed, but we have proposed an efficient and simple model which is described below.

The actual output data is the IT2F number $\tilde{y}_i = [[\underline{p}_i, \overline{p}_i], q_i, [\underline{r}_i, \overline{r}_i]]$. $\tilde{Y}_i = (\underline{\tilde{Y}_i}, \overline{\tilde{Y}_i})$ is the forecasted output and $\tilde{y}_i = (\underline{\tilde{y}_i}, \overline{\tilde{y}_i})$ is the actual output. Suppose the underlying thought of our method in line with Tanaka's is to minimize the fuzziness of the model by minimizing the fuzzy coefficients' total spread, subject including all the given data [9]. These can be satisfied, in T2FR, for primary and secondary membership functions.

We built our model according to the necessity model for LMF and the possibility model for UMF. These models can not only minimize the ambiguity in secondary membership function, but also contribute to finding a suitable model for fitting the predicted value to the observed one.

The followings are assumed for making a FLR model. An interval type-2 fuzzy linear (IT2FL) model shows the data in which $\tilde{Y}_i, \tilde{A}_j$ are entirely normal IT2FS and $X_j^{(i)}$ is positive:

$$\tilde{Y}_i = \tilde{A}_1 X_1^{(i)} + \cdots + \tilde{A}_q X_q^{(i)} = \sum_{j=1}^{q} \tilde{A}_j X_j^{(i)} \tag{1}$$

Thus, the supports of UMF and LMF are obtained as the following [31]:

$$\tilde{Y}_i = \left[\left[\underline{Ay_i}, \overline{Ay_i}\right], By_i, \left[\underline{Cy_i}, \overline{Cy_i}\right]\right] = \sum_{j=1}^{q} \tilde{A}_j X_j^{(i)}$$

$$= \sum_{j=1}^{q} \left[[\underline{a_j}, \overline{a_j}], b_j, [\underline{c_j}, \overline{c_j}]\right] X_j^{(i)} \tag{2}$$

$$= [\sum_{j=1}^{q} [\underline{a_j}, \overline{a_j}] X_j^{(i)}, \sum_{j=1}^{q} b_j X_j^{(i)}, \sum_{j=1}^{q} [\underline{c_j}, \overline{c_j}] X_j^{(i)}]$$

There for:

$$\underline{Ay_i} = \sum_{j=1}^{q} \underline{a_j} X_j^{(i)} \tag{3-1}$$

$$\overline{Ay_i} = \sum_{j=1}^{q} \overline{a_j} X_j^{(i)} \tag{3-2}$$

$$By_i = \sum_{j=1}^{q} b_j X_j^{(i)} \tag{3-3}$$

$$\underline{Cy_i} = \sum_{j=1}^{q} \underline{c_j} X_j^{(i)} \tag{3-4}$$

$$\overline{Cy_i} = \sum_{j=1}^{q} \overline{c_j} X_j^{(i)} \tag{3-5}$$

If $\tilde{Y}_i = \left[\left[\underline{Ay_i}, \overline{Ay_i}\right], By_i, \left[\underline{Cy_i}, \overline{Cy_i}\right]\right]$ and $\tilde{y}_i = [[\underline{p_i}, \overline{p_i}], q_i, [\underline{r_i}, \overline{r_i}]]$, the degree of the fitting of $\tilde{Y}_i$ to $\tilde{y}_i$ is measured by the h-level sets of the actual and the forecasted fuzzy numbers respectively, where h-level have been described according to definition 8. The vital notion is to acquire $\tilde{A}_j$, parameters $\underline{a_j}, \overline{a_j}, b_j, \underline{c_j}, \overline{c_j}$, such that:

$$\underline{\tilde{Y}_i} \subset_h \underline{\tilde{y}_i}, \qquad \overline{\tilde{y}_i} \subset_h \overline{\tilde{Y}_i} \tag{4}$$

We utilized the necessity for LMFs and the possibility model for UMFs. Since it is sufficient to close the membership functions of the observed and predicted values as much as possible, and an h-cut of the observed value is included in the predicted value [18].

The objective function is described by: $I = I_1 + I_2 - I_3 + I_4$

$$I_1 = \sum_{i=1}^{n} \sum_{j=1}^{q} [(\overline{a_j} - \underline{a_j}) X_j^{(i)}]^2 + \sum_{i=1}^{n} \sum_{j=1}^{q} [(\overline{c_j} - \underline{c_j}) X_j^{(i)}]^2 \tag{5-1}$$

$$I_2 = \sum_{i=1}^{n} \sum_{j=1}^{q} [(\overline{c_j} - \overline{a_j}) X_j^{(i)}]^2 \tag{5-2}$$

$$I_3 = \sum_{i=1}^{n}[q_i - \sum_{j=1}^{q} b_j X_j^{(i)}]^2 \qquad (5\text{-}3)$$

$$I_4 = \sum_{i=1}^{n}\sum_{j=1}^{q}[(\underline{c}_j - \underline{a}_j)X_j^{(i)}]^2 \qquad (5\text{-}3)$$

This four objective function work as follow:
1. The ambiguity of secondary membership function minimized by $I_1$.
2. The ambiguity of LMF of $\tilde{Y}$ that is a T1FS minimized by $I_2$.
3. The distance between point with the biggest membership rate in a predicted and the point with the highest membership value in the corresponding observed value minimized by $I_3$.
4. The necessity problem is met by $I_4$, hence it should be maximized.

The final goal is to find out the fuzzy parameters $\underline{a}_j, \overline{a}_j, b_j, \underline{c}_j, \overline{c}_j$, which are the solution of the following two QP problems and all objective function [18]:

$$Min\ I_2 \quad and\ Max\ I_3 \qquad (6)$$

s,t
$$By_i - (1-h)\left(By_i - \underline{Ay_i}\right) \le q_i - (1-h)\left(q_i - \underline{p}_i\right)$$
$$By_i + (1-h)(\overline{Cy_i} - By_i) \ge q_i + (1-h)(\overline{r}_i - q_i)$$
$$By_i - (1-h)(By_i - \overline{Ay_i}) \ge q_i - (1-h)(q_i - \overline{p}_i)$$
$$By_i + (1-h)\left(\underline{Cy_i} - By_i\right) \le q_i - (1-h)(\underline{r} - q_i)$$

Our problem can be rewrite as following:
$$Min\ I = I_1 + I_2 - I_3 + I_4 \qquad (7)$$

s,t
$$By_i - (1-h)\left(By_i - \underline{Ay_i}\right) \le q_i - (1-h)\left(q_i - \underline{p}_i\right)$$
$$By_i + (1-h)(\overline{Cy_i} - By_i) \ge q_i + (1-h)(\overline{r}_i - q_i)$$
$$By_i - (1-h)(By_i - \overline{Ay_i}) \ge q_i - (1-h)(q_i - \overline{p}_i)$$
$$By_i + (1-h)\left(\underline{Cy_i} - By_i\right) \le q_i - (1-h)(\underline{r} - q_i)$$
$$0 \le \underline{a}_j \le \overline{a}_j \le\ b_j \le \underline{c}_j \le\ \overline{c}_j$$

**3.2. General Type -2 Fuzzy Regression**

The suggested model is a developed Tanaka's model [9][26], whose underlying idea is to minimize the model's fuzziness by minimizing the general spread of fuzzy coefficients, subject to including all the provided data.

In the proposed model, vagueness in primary and secondary fuzzy sets should be minimized. Also, a specified h-plane of observed value should be included in the same h-plane of the predicted value.

Since h-plane build a FOU like area, therefore it should be discussed similar to interval type-2 fuzzy regression (IT2FR) model. Figure 4 shows h-cut that we employed in this paper. In this method, H-cut is employed in two stages: first in the secondary membership function and then in the primary membership function.

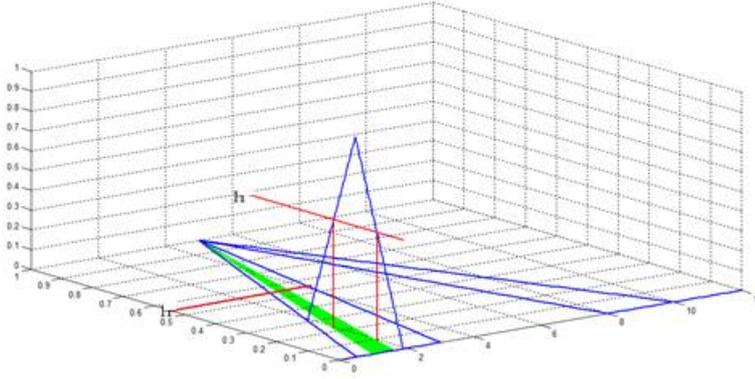

Figure 4. Illustration of h-cut of TT2FS.

After this h-plane representation, we calculate new parameter in new FOU. Figure 4 and 5 clearly shows support parameters for LMF and UMF of new FOU. Line equation of (1) and (2) that are legs of LMF and UMF are achievable (Figure 5). Hence, we can get 2 points ($y_1$ and $y_2$). With symmetric secondary membership function, B is given below:

$$B = \frac{y_1 + y_2}{2} \tag{8}$$

To find $x_{1h}$ and $x_{2h}$ that are support of new FOU, $y_{1h}$ and $y_{2h}$ should be find at first. The following relation can drive from secondary triangular membership function:

$$\frac{1}{1-h} = \frac{\alpha}{k} \Rightarrow k = \alpha(1-h) \tag{9}$$

Where: $\alpha = y_2 - B$
Thus:

$$y_{2h} = B + (1-h)(y_2 - B) \tag{10}$$

There for with two point $(q, 1)$ and $(x_l, y_{2h})$, 2h line equation can be derived and $x_{2h}$ has been found ($x_{2h}$ is the point that Y=0). Line equation can be written as:

$$Y - 1 = \frac{1 - y_{2h}}{q - x_l}(X - q) \tag{11}$$

If Y=0

$$x_{2h} = \frac{x_l - q}{1 - y_{2h}} + q \Rightarrow x_{2h} = \frac{x_l - q}{1 - B + (1-h)(y_2 - B)} + q \tag{12}$$

Similarly for $y_{1h}$:

$$y_{1h} = B - (1-h)(B - y_1)$$
$$\Rightarrow x_{1h} = \frac{x_l - q}{1 - (B - (1-h)(B - y_1))} + q \tag{13}$$

Therefore, right hand side can be obtained as follows:

$$x_{3h} = \frac{x_r - q}{1 - (B - (1-h)(B - y_3))} + q \tag{14-1}$$

$$x_{4h} = \frac{x_r - q}{1 - (B + (1-h)(y_4 - B))} + q \tag{14-2}$$

If the same calculation derived for predicted value, equations can be expressed as follow:

$$\hat{x}_{1h} = \frac{\hat{x}_l - By_i}{1 - (\hat{B} - (1-h)(\hat{B} - \hat{y}_1))} + By_i \tag{15-1}$$

$$\hat{x}_{2h} = \frac{\hat{x}_l - By_i}{1 - (\hat{B} - (1-h)(\hat{y}_2 - \hat{B}))} + By_i \tag{15-2}$$

$$\hat{x}_{3h} = \frac{\hat{x}_r - By_i}{1 - (\hat{B} - (1-h)(\hat{B} - \hat{y}_3))} + By_i \qquad (15\text{-}3)$$

$$\hat{x}_{4h} = \frac{\hat{x}_r - By_i}{1 - (\hat{B} - (1-h)(\hat{y}_4 - \hat{B}))} + By_i \qquad (15\text{-}4)$$

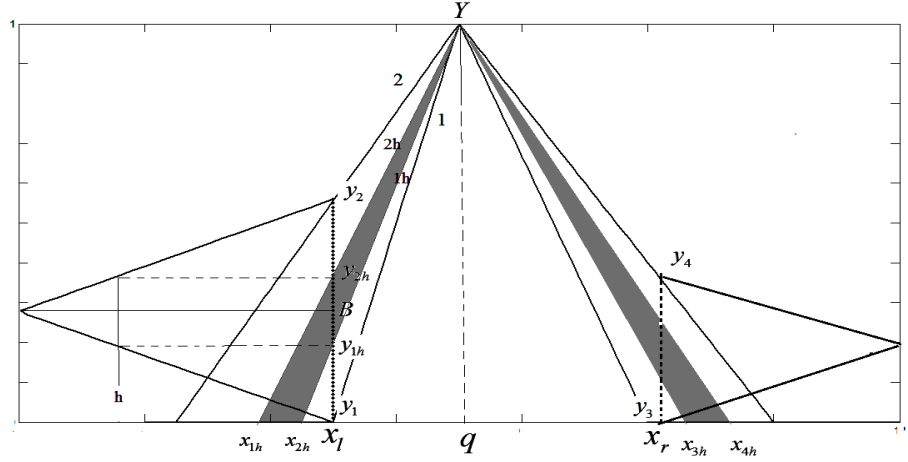

Figure 5. Membership function after applying h-cut.

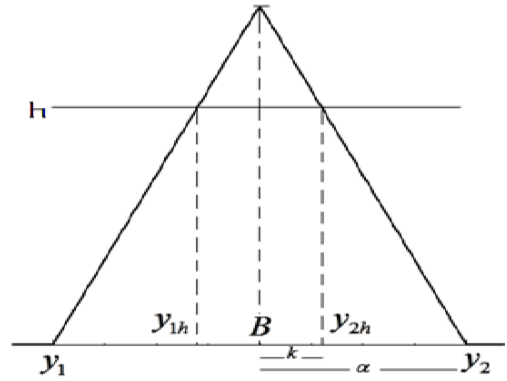

Figure 6. Secondary membership function in H-cut.

After calculation of $x_{1h}, x_{2h}, x_{3h}, x_{4h}$ and $\hat{x}_{1h}, \hat{x}_{2h}, \hat{x}_{3h}, \hat{x}_{4h}$ whereas we have an IT2FR problem that can be solved with previous IT2FR model. It is only enough to replace new parameters in IT2FR model. With replacement of $x_{1h}, x_{2h}, x_{3h}, x_{4h}$ with $\underline{p_i}, \overline{p_i}, q_i, \underline{r_i}, \overline{r_i}$ and $\hat{x}_{1h}, \hat{x}_{2h}, \hat{x}_{3h}, \hat{x}_{4h}$ with $\underline{Ay_i}, \overline{Ay_i}, By_i, \underline{Cy_i}, \overline{Cy_i}$, T2FR can be formulated as follow:

$$\text{Min } I = I_1 + I_2 - I_3 + I_4 \qquad (16)$$

s,t
$$By_i - (1-h)\left(By_i - \underline{Ay_i}\right) \le q_i - (1-h)(q_i - \hat{x}_{i1h})$$
$$By_i + (1-h)(\overline{Cy_i} - By_i) \ge q_i + (1-h)(\hat{x}_{i4h} - q_i)$$
$$By_i - (1-h)(By_i - \overline{Ay_i}) \ge q_i - (1-h)(q_i - \hat{x}_{i2h})$$
$$By_i + (1-h)\left(\underline{Cy_i} - By_i\right) \le q_i - (1-h)(\hat{x}_{i3h} - q_i)$$
$$0 \le \underline{a_j} \le \overline{a_j} \le \ b_j \le \underline{c_j} \le \ \overline{c_j}$$

## 4. Experimental Results

TAIEX forecasting data are utilized to measure the function of our suggested model. The TAIEX data in the year 2000 have been employed to show the ability of TT2FR model. Data from 11/2 to 12/15 are employed to train the model (training data). The data from 12/16 to 12/29 are test the performance of the model (testing data). H in TT2FR is adopted as 0.4 to obtain the optimal performance. Data are fuzzified into 58 TT2F sets based on main, low and high values obtained for many days. Model efficiency is evaluated by comparing RMSE and is given by [33][22][14-15]:

$$RMSE = \sqrt{\frac{\sum_{1=1}^{n}(actual\ value(t) - forcasted\ value(t))^2}{n}} \quad (17)$$

Here, $n$ is the number of data employed.

Fig. 7 and table 1 show results obtained using our proposed model. Figure 7 shows the results obtained using several T2FR. Qualitatively, our suggested model can predict the unknown data more closely. Table 1 represents the Summary of comparison with other state-of-art methods done using same database (TAIEX forecasting data). It can be noted from these two figures that, both for training and testing data our proposed TT2F model yielded more proper outcomes with minimum RMSE value as compared to other T2FR models. Our TT2FR model is able to handle the uncertainties in the data much better, hence it has yielded the least prediction error.

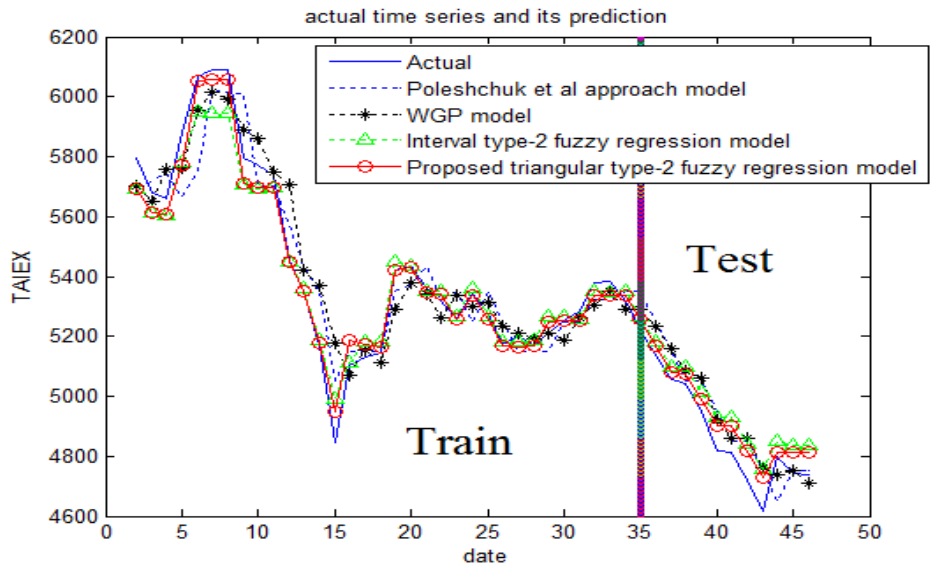

Figure 7. TAIEX forecasting results.

Figure 7 and Table 1 show that, the suggested model's root mean square error (RMSE) is inferior to RMSE of the model posited by Poleshchuk et al. [25], WGP model [27] and IT2FLR model [30]. Table 1 shows that, this model also performed better than the classic and fuzzy type 1 models. It shows the predictive capability of this suggested model is more effective than that of earlier T2FS models [15][34].

Due to the uncertainty in the nature of the problem data (like stock market) the forecasting method based on modeling uncertainty will be more useful. T2F models are more suitable to handle uncertainty as compared to other methods. Such models are optimized with different criteria to attain the maximum output with real data. Also, obtain less RMSE for both training and testing models. Our proposed model has obtained highest results and better accuracy in forecasting.

Due to the uncertainty in the nature of problem data (TAIEX), the forecasting method which is based on modeling uncertainty will have more favorable final answer. TT2FLR model develop the previous models from IT2FR model to triangular T2FR model, in addition, it is able to perform better in practice. In fact, the proposed model has shown its efficiency.

Table 1. Summary of comparison with other state-of-art methods done using same database (TAIEX forecasting data).

| RMSE Parameter | Classic Regression | Type-1 Fuzzy [34] | Type-1 Fuzzy Reg. [9] | Type-2 Fuzzy Models | | | | |
|---|---|---|---|---|---|---|---|---|
| | | | | Reference [35] | Reference [25] | Reference [27] | Reference [30] | Proposed Method |
| Train Data | 212 | 176 | 163 | 139 | 113 | 107 | 66 | 49 |
| Test Data | 282 | 189 | 201 | 165 | 182 | 179 | 109 | 95 |

Now, another example about COVID-19 detection in patients is explained. Because of high pandemic of this disease, better management of this disease across the world will be extremely useful. This disease is so infectious and there are many infected people all over the world. Consequently, if we are able to predict the number of potential patients more precisely, it will help us for designing better management systems to control COVID-19 pandemic [36-42].

The dataset used in this research was the number of patients in the world from January 22, 2020 till September 5, 2021 [43]. The results of applying the proposed method on these data were shown in Table 2.

Table 2. Summary of comparison with other state-of-art methods done using COVID-19 dataset [17]

| RMSE Parameter | Classic Regression | Type-1 Fuzzy [34] | Type-1 Fuzzy Reg. [9] | Type-2 Fuzzy Models | | | | |
|---|---|---|---|---|---|---|---|---|
| | | | | Reference [35] | Reference [25] | Reference [27] | Reference [30] | Proposed Method |
| Train Data | 8452 | 6923 | 7245 | 4356 | 3934 | 3610 | 3167 | 2960 |
| Test Data | 9592 | 7210 | 7012 | 5583 | 4567 | 4795 | 3845 | 3079 |

In Figure 8, the mortality prediction in COVID-19 patients was plotted.

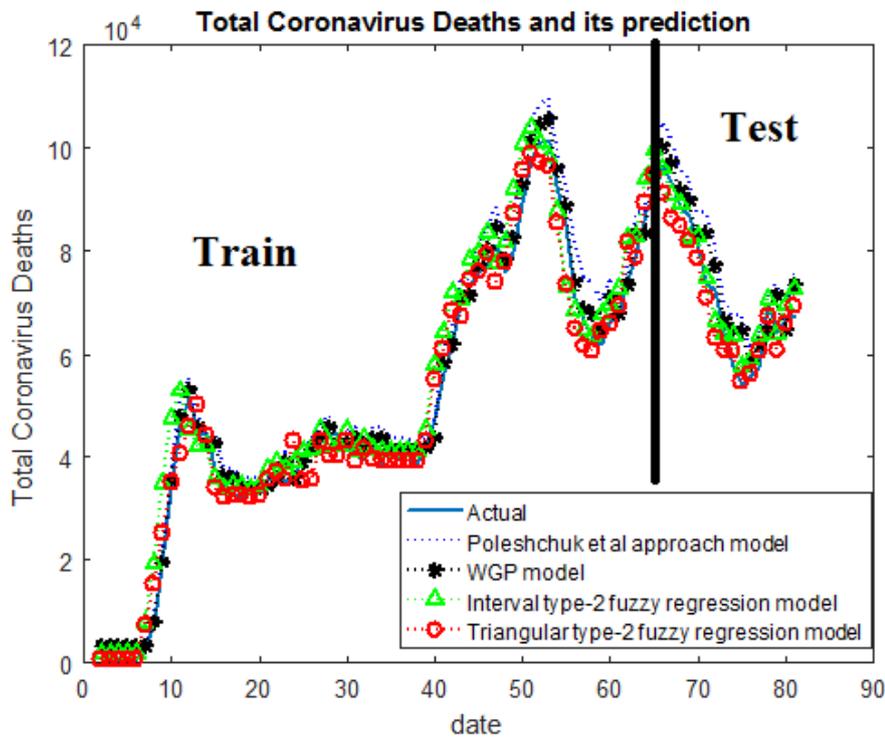

Figure 8, daily prediction of world mortality in COVID-19 patients

According to Figure 8, it is clear that the proposed method had acceptable performance in prediction of mortality in COVID-19 patients.

The advantages of the proposed model are as follows:
(i) Proposed model is simple and does not require more computation.
(ii) Able to overcome the uncertainty efficiently and predict accurately.
(iii) Model is robust even in noisy environment.
(iv) Such models can also be used in other uncertain prediction applications such as stock market prediction etc.

Limitations of our work are as follows:
(i) Model has been developed using only one database (TAIEX forecasting data). It needs to be tested with more data in future.

(ii) The value of H is considered by trial and error while it can be find by an optimization approach.

## 5. Conclusions and Future Works

The current survey shows a triangular regression model of T2F in an efficient and sensible way. The proposed model IT2FS model is less computationally intense to use due to its smaller dimension. Also, this model can easily extend other types of membership functions. This model helps to model more uncertainty without adding additional complexity. Such models encourage researchers to use T2F models without the overheads of its 3D form. For the evaluation of our model, TAIEX and COVID-19 forecasting is employed, which is a commonly employed benchmark to show model efficiency. From the empirical results, it is inferable that this model can adequately forecast the Taiwan stock index and COVD-19 in addition to its intelligible steps; therefore, it can be employed in tasks with considerable uncertainty. For future works, one of the best paths to investigate is changing the inputs, such as developing a TT2FLR model with T2F input instead of crisp input. This change in input may help in modeling uncertainty, and also can improve results.

Additionally, as the future work, it is possible to apply the proposed method for detection of various diseases such as epileptic seizure [44-46], Autism spectrum disorder (ASD) [47], schizophrenia [48-49], Multiple Sclerosis (MS) [50], and heart diseases [51].

**Funding:** This research received no external funding.

**Conflicts of Interest:** The authors declare no conflict of interest.